\documentclass[a4paper, 10pt, conference]{ieeeconf}      

\IEEEoverridecommandlockouts                              

\usepackage{soul}
\usepackage{color}
\usepackage[binary-units=true]{siunitx}
\usepackage{graphicx}
\usepackage{gensymb}
\usepackage{hyperref}
\usepackage[caption=false, font=footnotesize,position=top]{subfig}
\usepackage{tabularx, booktabs}
\usepackage{multirow}
\usepackage[utf8]{inputenc}

\title{\LARGE \bf
Towards Autonomous Driving: a Multi-Modal 360\degree\ Perception Proposal
}

\author{Jorge Beltrán$^{1}$, Carlos Guindel$^{1}$, Irene Cortés$^{1}$, Alejandro Barrera$^{1}$, Armando Astudillo$^{1}$, Jesús Urdiales$^{1}$,\\ Mario Álvarez$^{1}$, Farid Bekka$^{2,\dagger}$, Vicente Milanés$^{2,\dagger}$ and Fernando García$^{1,*}$
\thanks{$^{1}$J. Beltrán, C. Guindel, I. Cortés, A. Barrera, A. Astudillo, J. Urdiales, M. Álvarez and F. García are with the Intelligent Systems Lab, Universidad Carlos III de Madrid, Leganés, Spain.}
\thanks{$^{2}$V. Milanés and F. Bekka  are with the Research Department, Renault SAS, Guyancourt, France.}
\thanks{$^{\dagger}$This paper reflects solely the views of the authors and not necessarily the views of the company they belong to.}
\thanks{$^{*}$Correspondence author: {\tt\small fegarcia@ing.uc3m.es}}}

\begin{document}
\bstctlcite{IEEEexample:BSTcontrol}

\maketitle
\thispagestyle{empty}
\pagestyle{empty}

\begin{abstract}

In this paper, a multi-modal 360\degree\ framework for 3D object detection and tracking for autonomous vehicles is presented. The process is divided into four main stages. First, images are fed into a CNN network to obtain instance segmentation of the surrounding road participants. Second, LiDAR-to-image association is performed for the estimated mask proposals. Then, the isolated points of every object are processed by a PointNet ensemble to compute their corresponding 3D bounding boxes and poses. Lastly, a tracking stage based on Unscented Kalman Filter is used to track the agents along time. The solution, based on a novel sensor fusion configuration, provides accurate and reliable road environment detection. A wide variety of tests of the system, deployed in an autonomous vehicle, have successfully assessed the suitability of the proposed perception stack in a real autonomous driving application.


\end{abstract}

\section{INTRODUCTION}

Transport is one of the main drivers of economic development and improved accessibility in modern societies. However, most modes used today have numerous adverse effects, posing a significant challenge for future sustainable development. Of particular concern is the case of road transport, which suffers from a variety of problems such as traffic crashes, one of the leading causes of death in the world, and greenhouse gas emissions, responsible for global warming.

Autonomous vehicles, together with alternative powertrains (e.g., electric motors), offer a response to these challenges that will represent a paradigm shift in mobility in the coming decades. Research in autonomous driving dates back a few decades and has led to significant advances in areas such as automatic vehicle control and environment perception. However, there is still a critical technical issue that must be solved before moving on to industrial-level considerations: these systems must be able to fulfill the role of the human driver by closing the vehicle control loop, which involves a robust perception of the environment to support the decision making procedure.

One of the central tasks of the perception subsystem is the detection of the rest of road users in the surroundings of the vehicle, given that this function cannot rely solely on a~priori information. This function provides proper awareness of the dynamic traffic situation and is therefore essential for autonomous navigation. Data to perform this task comes from exteroceptive sensors, on which the accuracy of the final result largely depends. In this regard, the combined use of different modalities offers increased robustness and redundancy. 

\begin{figure}
    \centering
    \includegraphics[width=\columnwidth]{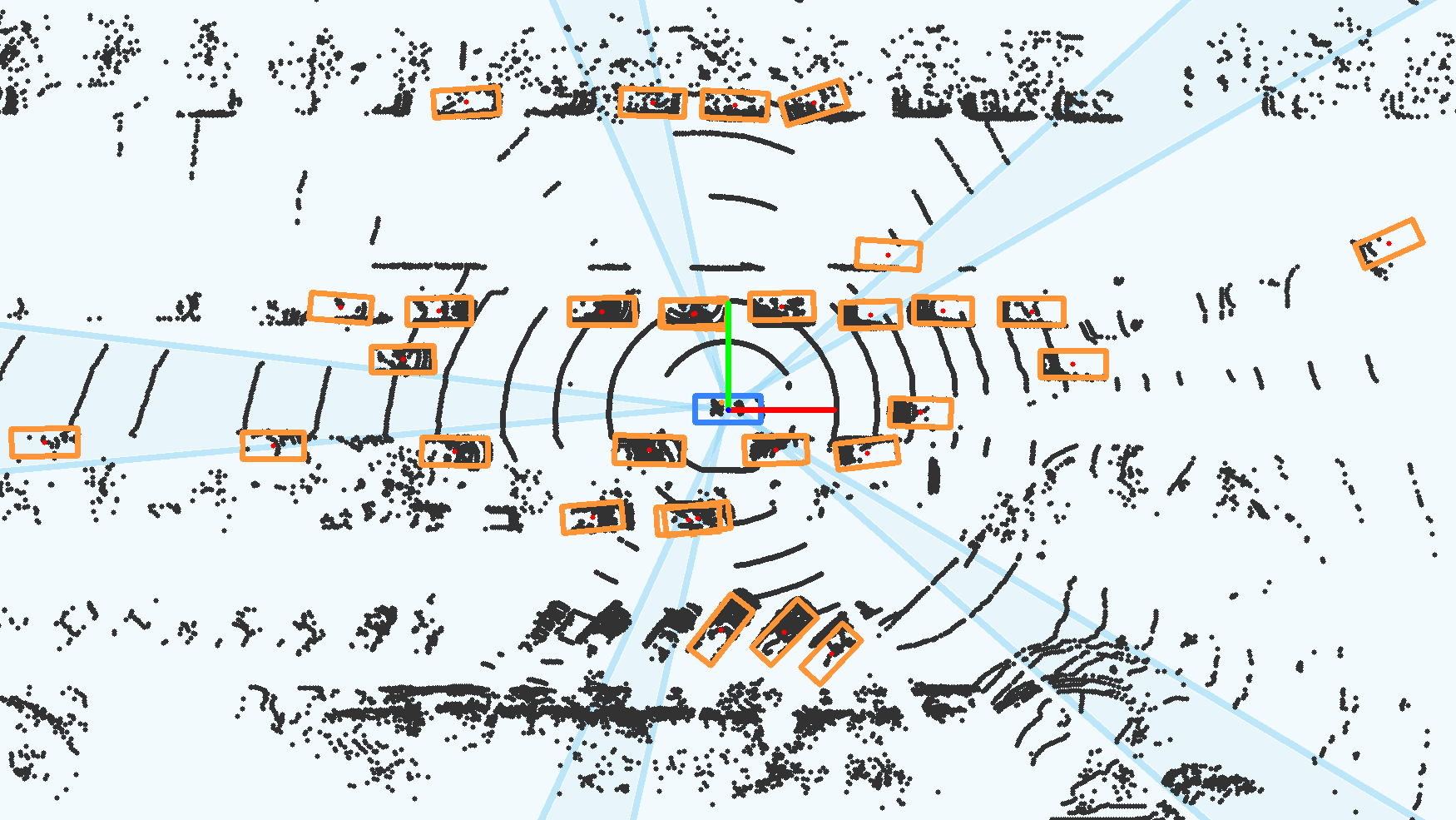}
    \caption{Bird's Eye View representation of the results provided by the proposed pipeline, with the ego-car placed in the center of the frame. The aggregated field of view of the cameras covers 360\degree. Darker blue represents the overlapping areas between cameras.}
    \label{fig:fancy}
\end{figure}

We propose a solution for 360\degree\ perception of the environment of an autonomous vehicle based on visual and LiDAR information. The approach, based on cutting-edge deep-learning-based methods, is aimed to perform a detailed identification of road users around the vehicle, including properties such as their classification, position, geometry, and speed. As shown in Fig.~\ref{fig:fancy}, this information allows delivering a precise environment model to the vehicle's high-level decision modules. The integration of the proposed system into a fully-robotized Renault ZOE prototype has confirmed that this outcome fulfills the requirements for autonomous navigation in a wide range of shared traffic situations.




\section{RELATED WORK}
\subsection{Sensor Setup}
Among the different modalities that are commonly used for onboard perception, cameras and LiDAR are a common choice in autonomous vehicle research \cite{Tas2018, Heng2018}. The underlying reason is the complementarity of their inherent properties: cameras can provide detailed information about the environment, which is useful for tasks such as classification, whereas LiDAR scanners retrieve accurate information about the spatial configuration of the environment within a wide operative range \cite{VanBrummelen2018}. Although the most critical area is the one right in front of the car, proper scene awareness requires extending the capabilities to cover the complete 360\degree\ range around the vehicle. Modern multi-layer LiDAR scanners suit this requirement naturally; however, the limited Horizontal Field of View (HFOV) of cameras makes it necessary to include an array of devices looking at different locations, as shown in recent traffic-oriented datasets such as nuScenes \cite{Caesar2019} or Waymo Open Dataset \cite{Sun2019}.

\subsection{3D Object Detection}
3D object detection encompasses the detection, classification, and positioning of obstacles within the FOV of the sensors. While this function can be performed by using data from a single modality, either images \cite{Ma2019} or LiDAR scans \cite{Yan2018}, the fusion of both sources during inference allows exploiting their respective strengths for reasoning. Current approaches aimed at 3D object detection via data fusion can be divided into two main groups. On the one hand, some proposals embed fusion within the CNN-based inference procedure itself, e.g., MV3D \cite{Chen2016b} or AVOD \cite{Ku2018}. On the other hand, an alternative two-stage approach is used by some methods such as F-PointNet \cite{Qi2017a} and F-ConvNet \cite{Wang2019b}. These methods take advantage of a pre-computed set of detections in the image and output a 3D cuboid representing each one from LiDAR information. In that way, the final result benefits from the high accuracy provided by modern object detectors in images, such as RetinaNet \cite{Lin2017} or Faster R-CNN \cite{Ren2015a}, while also providing an accurate estimate of the geometrical properties of the objects.

\subsection{Tracking}
Typically, object detection is based on a single data frame and, therefore, provides an instantaneous view of the environment. Conversely, the tracking stage allows maintaining an estimate of the state of the dynamic agents over time, thus providing temporal consistency. Most tracking approaches for onboard perception are based on particle \cite{cho2006real} or Kalman \cite{Garcia2017} filters. Non-linear variants of the latter, such as the Unscented Kalman Filter (UKF) \cite{wan2000unscented}, are the most popular because they can represent better the movement of the road users. These methods require a data association stage to match detections from different frames, e.g., the Hungarian algorithm \cite{Sharma2018}. Recent trends in object tracking try to exploit deep learning structures, such as Recurrent Neural Networks (RNNs) \cite{milan2017online} or siamese networks \cite{schroff2015facenet}, to improve the performance of the tracking stage. 

\section{SENSOR CONFIGURATION} 
This section describes the sensor setup of the proposed 360\degree\ perception solution. The following sensors are employed:
\begin{itemize}
\item Five CMOS cameras equipped with an 85\degree-HFOV lens. 
\item A 32-layer LiDAR scanner featuring a minimum vertical resolution of 0.33\degree\ and a range of \SI{200}{\meter} (Velodyne Ultra Puck).
\end{itemize}
All the sensors are mounted on a rack on the roof of the vehicle. The LiDAR scanner stands in a central position, while the cameras are evenly distributed around it to cover a full 360\degree\ HFOV around the vehicle with some overlap between images. 
Sensors have been selected to ensure optimal performance in the short-to-medium range. The image resolution provided by the cameras' sensors is high enough to enable the use of pixel binning (by a factor of $2\times$ in both directions) to increase the sensitivity to light. 


This configuration has been chosen to exploit the benefits of using both sensor modalities together. With this goal in mind, accurate synchronization and calibration between sensors are of paramount importance. Hence, they all are synchronized with the clock provided by a GPS receiver, and cameras are externally triggered at a \SI{10}{\hertz} rate. Regarding calibration, cameras' intrinsic parameters are obtained through the checkerboard-based approach by Zhang \cite{Zhang2000}, and extrinsic parameters representing the relative position between sensors are estimated through a monocular-ready variant of the \textit{velo2cam} method introduced in \cite{Guindel2017ITSC}. The result of this automatic procedure is further validated by visual inspection.



\section{SOFTWARE ARCHITECTURE}

\begin{figure*}
    \centering
    \includegraphics[width=\textwidth]{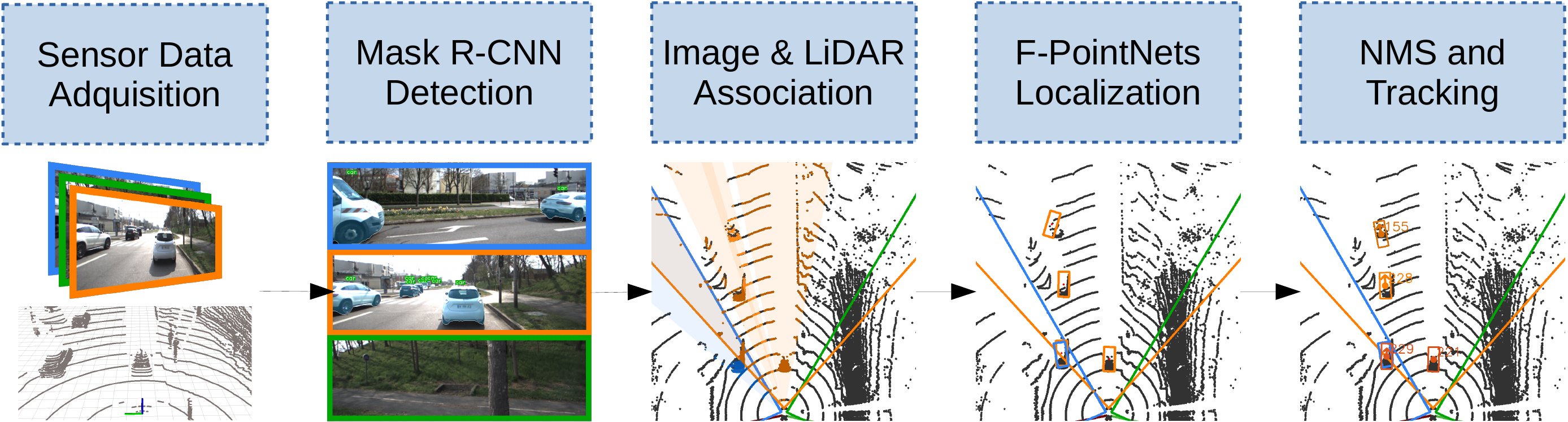}\
    \caption{System overview. Images from all the cameras are processed by individual instances of Mask R-CNN, which provide detections endowed with a semantic mask. LiDAR points in these regions are used as an input for several F-PointNets responsible for estimating a 3D bounding box and estimate its position with respect to the car. Then, 3D detections from each camera are fused using an NMS procedure. A subsequent tracking stage provides consistency across several frames and avoids temporary misdetections.}
    \label{fig:overview}
\end{figure*}
The proposed solution is based on three pillars. First, visual data is employed to perform detection and instance-level semantic segmentation. Then, LiDAR points whose image projection falls within each obstacle bounding polygon are employed to estimate its 3D pose. Finally, the tracking stage provides consistency, thus mitigating occasional misdetections and enabling trajectory prediction. The combination of these three stages allows accurate and robust identification of the dynamic agents surrounding the vehicle. 


The following sections are devoted to describing each of these parts in detail. The software architecture is fully implemented on ROS\footnote{\url{https://www.ros.org/}} and is endowed with a high-level supervisor able to control the status of every node and detect critical malfunctions while in operation.


\subsection{Vision-based 2D Detection}
Vision-based sensors (i.e.,~cameras) provide high-resolution appearance features that provide superior performance over other sensor modalities for the task of object detection. Because of that, our solution relies on a computer-vision-based solution at the early stages of the pipeline. In particular, we adopt a two-stage detector with instance segmentation capabilities, Mask R-CNN \cite{He2017}. This framework is an extension of the well-known Faster R-CNN approach \cite{Ren2015a} that adds a new branch responsible for estimating a pixel-wise mask per object. 

As has been proven in the literature \cite{Huang2017}, two-stage methods such as Faster R-CNN offer better performance in the detection of small objects than their single-stage counterparts. This feature becomes critical in driving applications. Although they are also generally slower, modern implementations achieve framerates that are compatible with the requirements of online processing. 

The improvement of Mask R-CNN over Faster R-CNN is twofold: not only it provides an additional semantic mask for each object but also enhances the detection performance due to the benefits of adding pixel-wise annotations into the multi-task training procedure \cite{He2017}. Besides, the computational burden added by the mask estimation branch is not very significant, as features are still computed only once. 

This detection paradigm divides the detection inference into two stages that make use of a shared set of convolutional features. At the first stage, a lightweight Region Proposal Network (RPN) propose regions where an object is likely to be found. Then, the second stage processes separately each proposal by extracting the corresponding features and feeding them to three individual branches or \textit{heads} that provide an estimate of the category, bounding box refinement, and semantic mask for each instance. All these tasks make use of the features computed by a feature extractor or \textit{backbone}. The whole pipeline is trained jointly through a multi-task loss that takes into account the contribution of the different outputs of the RPN and the three per-instance outcomes.

In our proposal, five different instances of Mask R-CNN, one per camera, are run in parallel. The inference is performed on a region of interest cropped from the original image that does not include those areas belonging to the own vehicle (e.g., the hood) and the sky. We use the ResNet-50 \cite{He2015b} as a backbone due to its excellent accuracy/speed tradeoff. Additionally, as small objects are one of the main focus of our system, we adopt the FPN (Feature Pyramid Network) approach \cite{Lin2016} to generate additional feature maps at different scales, that are subsequently used by the classification stage to sample features through the usual \textit{ROIAlign} layer. 

The Cityscapes dataset \cite{Cordts2016a} is employed to train Mask R-CNN, although we use a two-stage pretraining with makes use of ImageNet \cite{Russakovsky2015} and COCO \cite{Lin2014}, one after another, to enhance the quality of the convolutional features. Besides, we combine each instance of \textit{bicycle} with its closer \textit{rider} to obtain a \textit{Cyclist} annotation, according to the KITTI dataset \cite{Geiger2012b} labeling policy, which will be used down the pipeline.

\subsection{3D Bounding Box Estimation}
Regardless of the excellent detection performance of Mask R-CNN, there is still a gap between the 2D detections, expressed in image coordinates, and the desired outcome of the proposed perception system, which is the 3D location and pose of every detected agent. LiDAR information is included here to perform spatial reasoning. Given the extrinsic parameters relating each camera with the LiDAR scanner, the correspondence between detections in image and LiDAR points (resulting in RGB-D data) is straightforward; however, the estimation of 3D bounding boxes representing the geometry and position of the objects is far from a trivial task since LiDAR data only represent the non-occluded surface of the objects.

We adopt the Frustrum PointNet (F-PointNet) approach \cite{Qi2017a} to perform the necessary inference step and deliver an accurate estimate of the location, size, and orientation of obstacles based on raw LiDAR data. F-PointNet accepts as an input the LiDAR points belonging to an obstacle; therefore, it relies upon a set of pre-computed detections to perform the assignment between points and obstacles. In the original approach, each obstacle is assigned the points that fall within the frustum extruded from the image detection. 

The idea behind F-PointNet leads to optimal leverage of visual and LiDAR information. On the one hand, the whole 3D detection process is based on the result of the image inference, which offers excellent performance in detecting road users. On the other hand, it also exploits the highly accurate and precise geometric information given by the LiDAR modality in those tasks in which it is most appropriate. Besides, the method achieves outstanding results, among the best in renowned benchmarks such as KITTI, while using very lightweight models for inference.

We use the F-PointNet v1 model, as the enhancement introduced by the v2 version was proved not significant in this context. This architecture is made of three sequential stages that lead to the final 3D box estimation. The first stage is a 3D instance segmentation PointNet that performs point-wise binary classification to filter out those points contained in the frustum that do not represent the object itself (i.e.,~noise). The second stage consists of a light network (T-Net) responsible for estimating the center of the object, which allows normalizing the position of all points around this reference. Finally, an amodal 3D box estimation PointNet is employed to infer each 3D box coordinate through a hybrid of classification and regression. 

In the proposed solution, as with the 2D detection, the inference is performed individually for each of the cameras. As a consequence of the use of Mask R-CNN in the detection stage, we feed the F-PointNet pipeline with those points that correspond to the instance segmentation masks, instead of the full 2D bounding box.
Due to the large number of LiDAR measures available, the association between masks and LiDAR points is performed through a GPU-based procedure. We still employ the original segmentation network to avoid minor errors caused by both the difference in perspective between both sensors and eventual miscalibrations.

The KITTI dataset \cite{Geiger2012b} is used to train the model. The data provided by the LiDAR scanner in our sensor setup is substantially different from the one featured by KITTI due to differences in the number of layers and their distribution. In order for the generated model to have an adequate representation of the input data to be used in inference, we use the method introduced in \cite{Beltran2019} to synthetically generate the training LiDAR data from the real KITTI frames, according to the particularities of our custom sensor setup. 


At the end of this stage, we have a set of 3D cuboids representing the objects in the FOV of each camera. They are subsequently expressed in a common coordinate frame, which is that of the LiDAR. Objects in the overlapping areas of contiguous cameras cause some instances of duplicated detections; for that reason, we apply a Non-Maximum Suppression (NMS) procedure on Bird's Eye View (BEV) (i.e.,~ignoring the height coordinate) to those detections located in those areas. For efficiency reasons, a per-class axis-aligned NMS approach is used.

\subsection{Tracking}
Once the full set of detections is available, the tracking stage comes into play. This task complements the previous stages by adding consistency to the detections and predicting the current position of obstacles even when they are not visible for a limited time (e.g.,~due to occlusions). The tracking algorithm in use in this proposal makes use of the square root version of the Unscented Kalman Filter (UKF) \cite{wan2000unscented} to estimate the dynamic state of each detected obstacle. The use of this variant of the UKF brings extra stability to the filter since it helps to avoid numerical errors \cite{Merwe2001}.

In the proposed solution, each obstacle type has a different system model, so that the prediction is adapted to the behavior of every agent. Taking advantage of the information retrieved by the previous stages of the pipeline, state variables are related to the 3D pose of the obstacles, and the filter is initially fed with the rotation estimation of the bounding box. As the sensors are mounted on a moving platform, the movement of the ego-vehicle is compensated using location information from a GPS  receiver and orientation information from an IMU.

On the other hand, the association of the detections between frames is performed using the Hungarian algorithm, where the costs of association between predictions and detections are computed through the Mahalanobis distance to include the uncertainty estimated by the Kalman filter. The track management allows objects to be tracked even during several frames after they cannot be associated with any detection, so the method is robust to occlusion or misdetections.


\section{RESULTS}

The proposed solution has been deployed on an automated vehicle and validated in a variety of traffic situations. The set of tests lasts several hours of operation and comprises a manifold of urban and peri-urban scenarios with different road users. Although all the stages of the pipeline rely on well-proven approaches, we present here a systematic analysis to assess the adequacy of the full system, based on a combination of both controlled experiments and real scenarios. 

\subsection{Quantitative Analysis} 

\begin{figure*}
\captionsetup[subfloat]{farskip=2pt,captionskip=1pt}
\subfloat{%
    \includegraphics[scale=0.48]{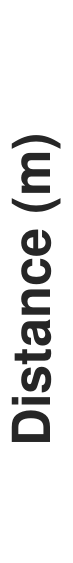}
}
\setcounter{subfigure}{0}
\hspace{-6px}
\subfloat[Sequence 1]{%
    \includegraphics[scale=0.48]{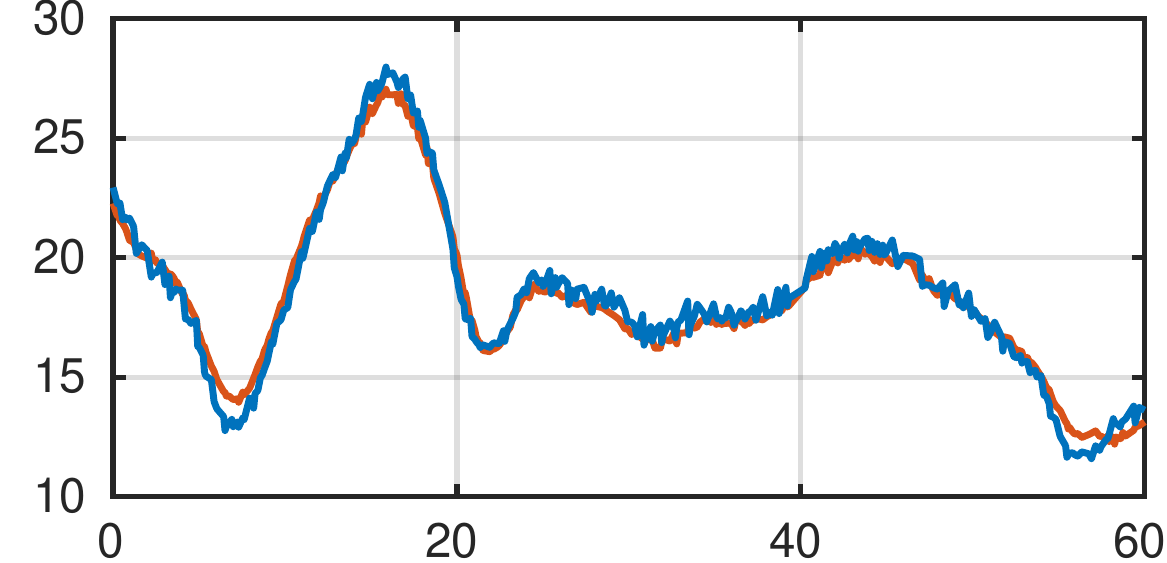}
}
\subfloat[Sequence 2]{
    \includegraphics[scale=0.48]{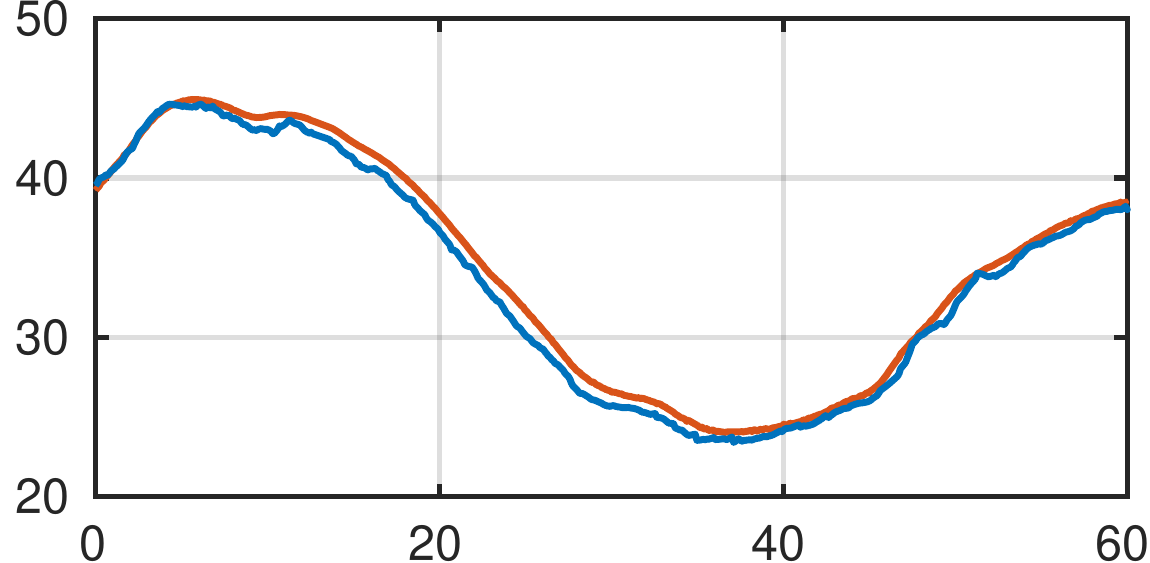}
}
\subfloat[Sequence 3]{
    \includegraphics[scale=0.48]{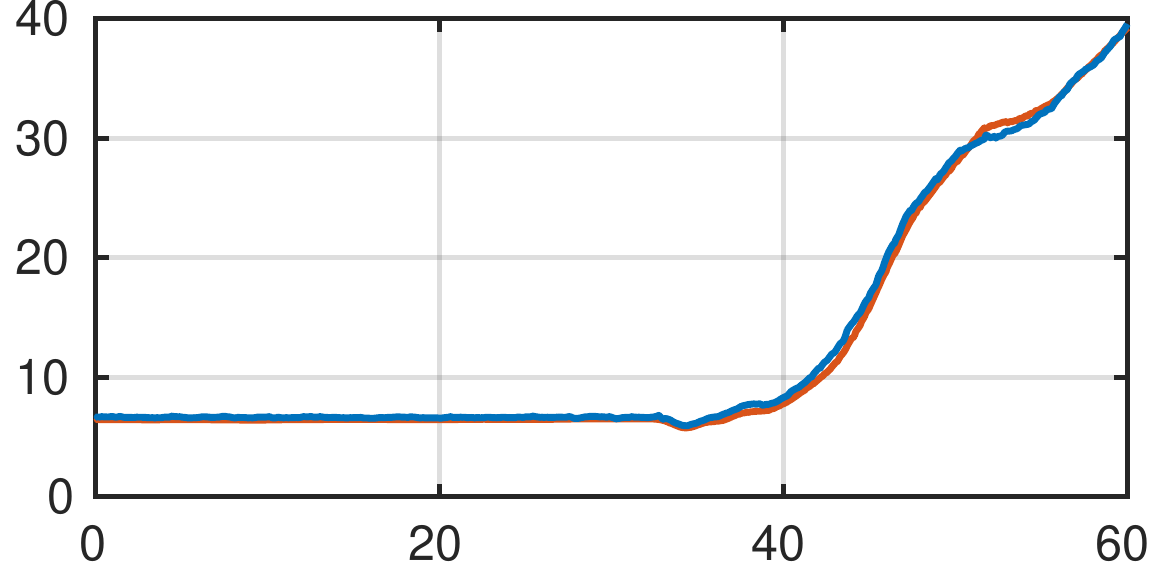}
} \\ 
\subfloat{
\\
    \includegraphics[scale=0.48]{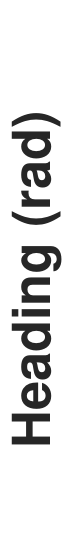}
}
\setcounter{subfigure}{0}
\hspace{-6px}
\subfloat{
\\
    \includegraphics[scale=0.48]{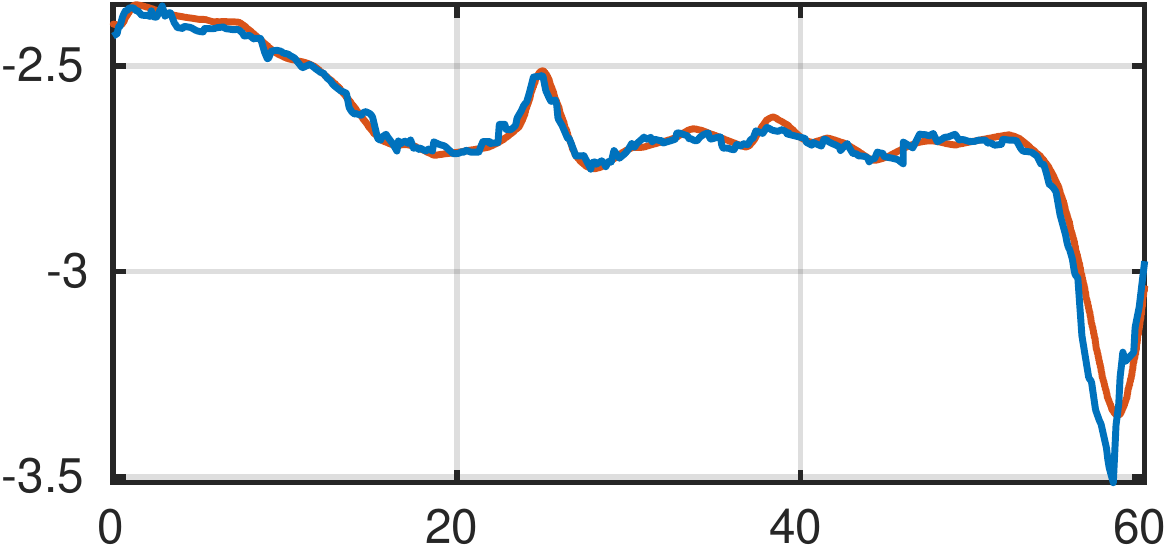}
}
\subfloat{
    \includegraphics[scale=0.48]{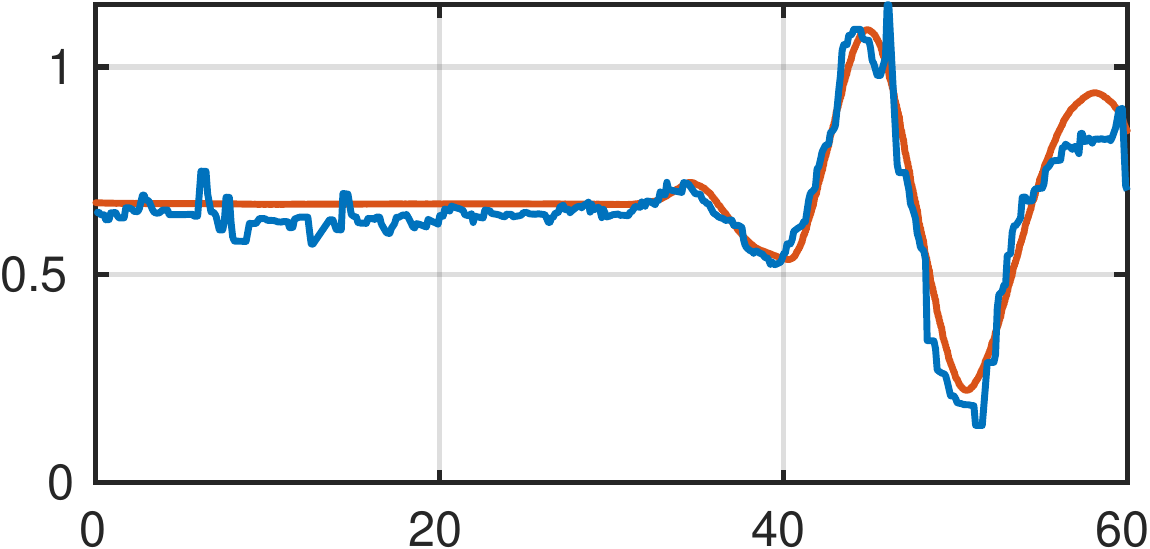}
}
\subfloat{
    \includegraphics[scale=0.48]{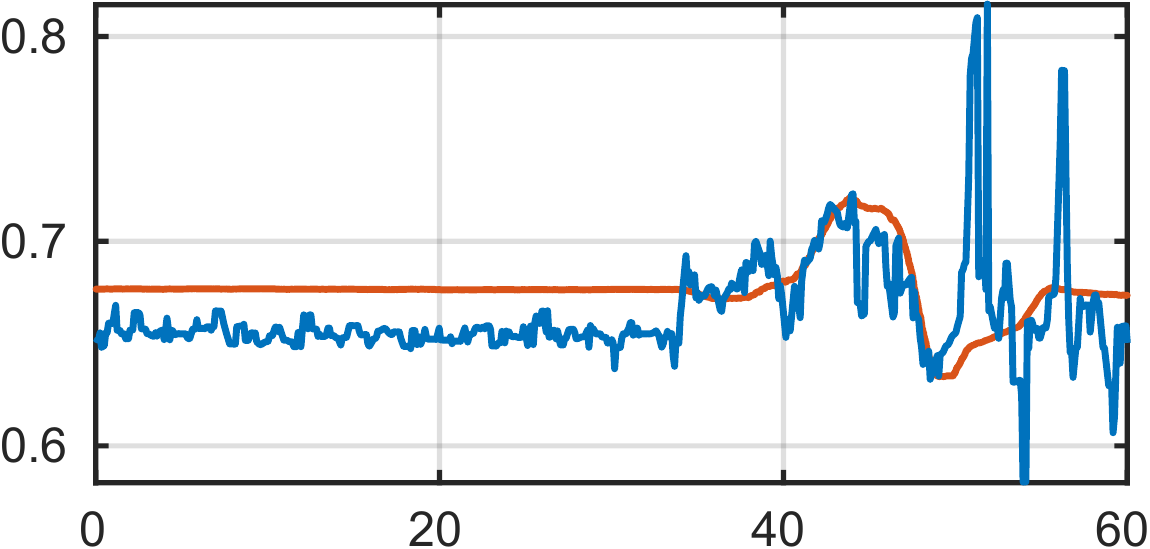}
} \\ 
\subfloat{
\\
    \includegraphics[scale=0.48]{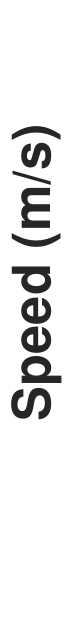}
}
\setcounter{subfigure}{0}
\hspace{-6px}
\subfloat{
\\
    \includegraphics[scale=0.48]{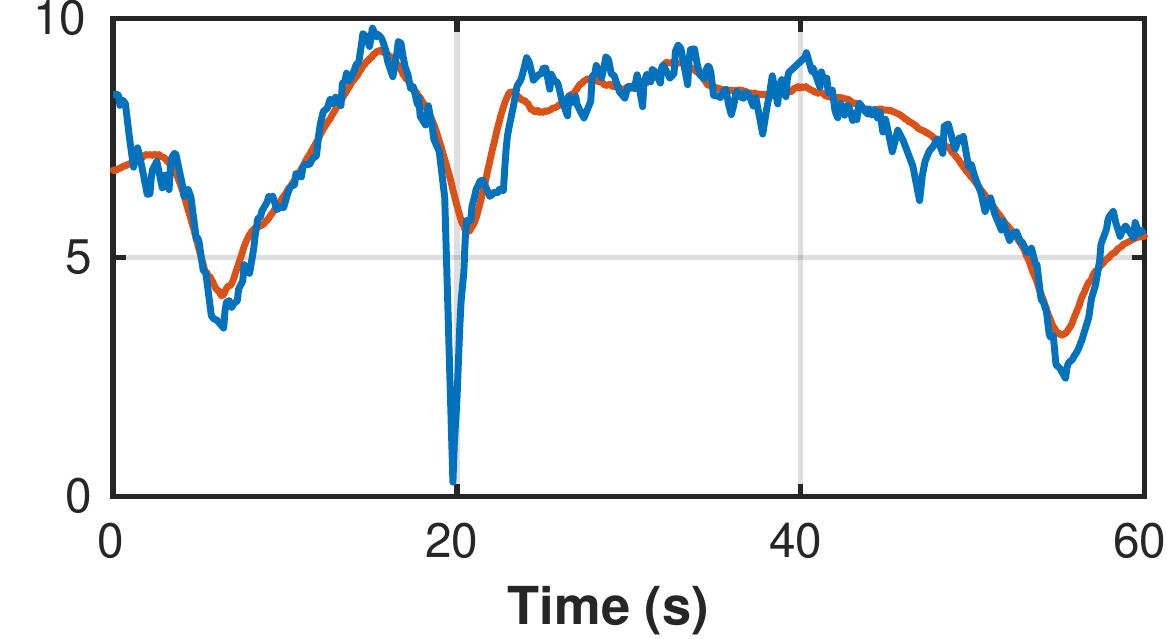}
}
\subfloat{
    \includegraphics[scale=0.48]{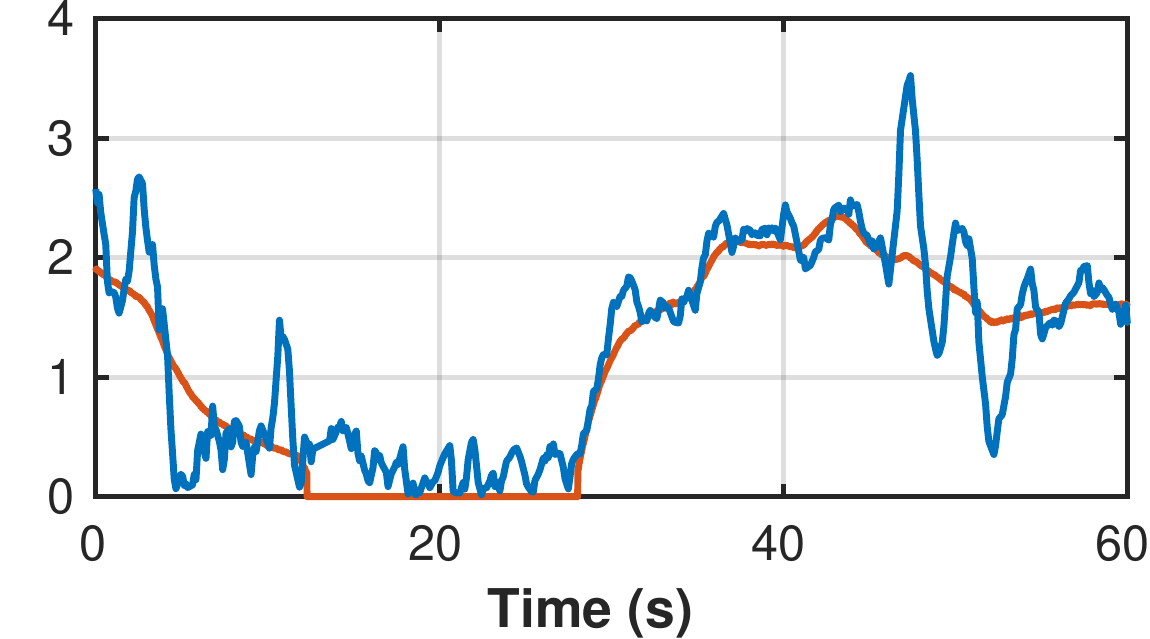}
}
\subfloat{
    \includegraphics[scale=0.48]{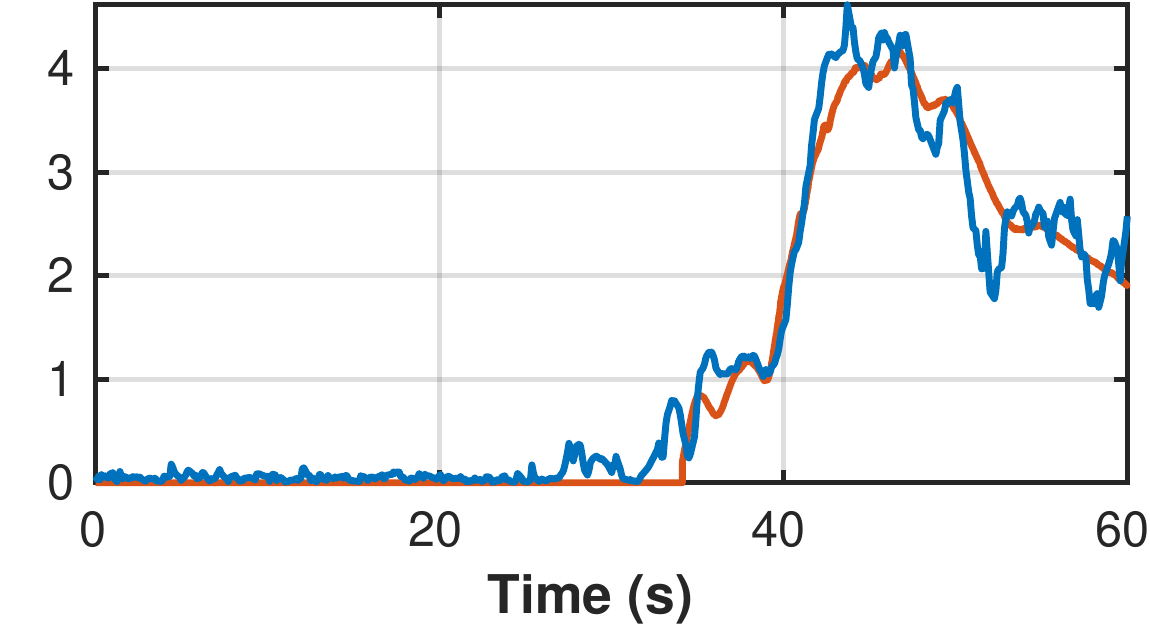}
}
\caption{Results obtained on three testing sequences used for assessing the performance of the approach. Graphs show the comparison between the output of the system (distance, heading, and velocity), in blue, and the GPS+RTK correction used as ground truth, in orange.}
\label{fig:graphs}
\end{figure*}

First, we assess the suitability of the method in a setup with an instrumented vehicle driving ahead. This leading vehicle is endowed with a GPS+IMU unit with RTK correction so that its position, orientation, and speed are known, which enables the evaluation of the proposed detection solution. The recorded sequences involve other road agents, but due to the lack of ground-truth annotations, the presented analysis is limited to the detection and tracking of the reference vehicle. Fig.~\ref{fig:graphs} shows the results of three typical use cases: an urban roundabout with moving traffic, a lane change during a traffic jam, and a pedestrian crossing.

As can be seen, the system is able to detect and track the reference vehicle with high accuracy, providing minimal error in distance and heading estimation. Remarkably, the performance of the perception pipeline remains stable beyond \SI{40}{\meter}. Regarding the speed, although the predicted output is generally noisier, the error in the estimation remains reasonably low. It is noteworthy that the filter is robust against the inaccuracy of GPS measurements in urban areas, which might affect the ego-vehicle movement compensation stage.


Table~\ref{tab:seq_graph} shows the average estimation errors of the pipeline on the above sequences. The metrics include the mean distance (on the ground plane), mean heading, and mean speed errors per sequence. The results observed on the different tested scenarios prove the high reliability and accuracy of the approach and demonstrate its proficiency on consistently tracking agents along time.

\begin{table}[htb] 
	\caption{Results for the three test sequences included in Fig. \ref{fig:graphs}}
	\label{tab:seq_graph}
	\centering
	\begin{tabular}{l c c c  }
		\toprule
		 & Seq. 1 & Seq. 2 & Seq. 3 \\ 
		 \midrule   
		 Mean distance error (m) & 0.48 & 0.71 & 0.30   \\ 
         Mean heading error (rad) & 0.02 & 0.04 & 0.02 \\
         Mean speed error (m/s) & 0.42 & 0.27 & 0.17   \\ 
		\bottomrule
	\end{tabular}
\end{table}       

\subsection{Qualitative Results in Real Traffic Scenarios}
\begin{figure*}
\captionsetup[subfloat]{farskip=2pt,captionskip=1pt}
\subfloat{
\\
    \includegraphics[width=0.16\linewidth]{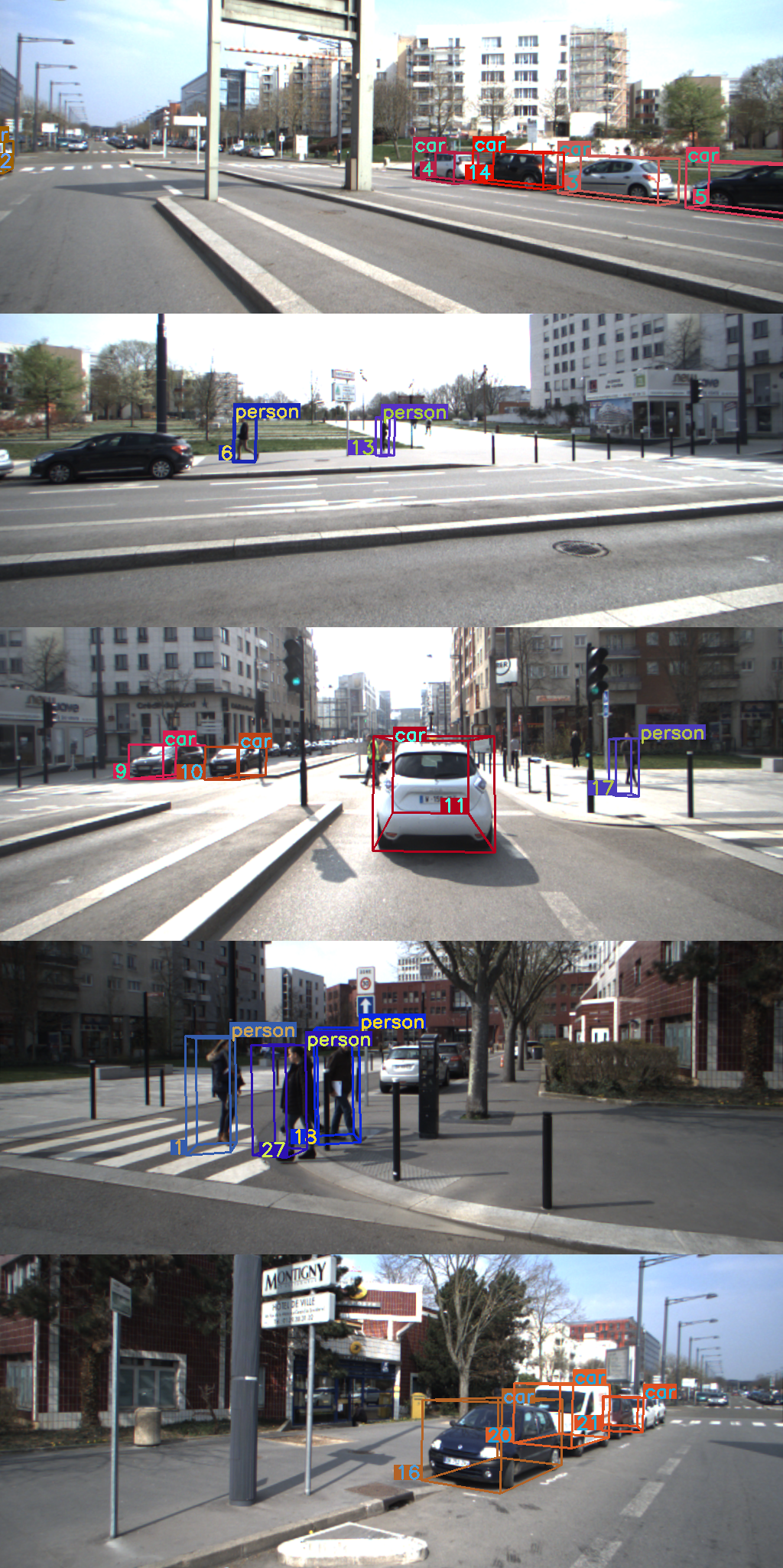}\hspace{0.7px}
    \includegraphics[width=0.16\linewidth]{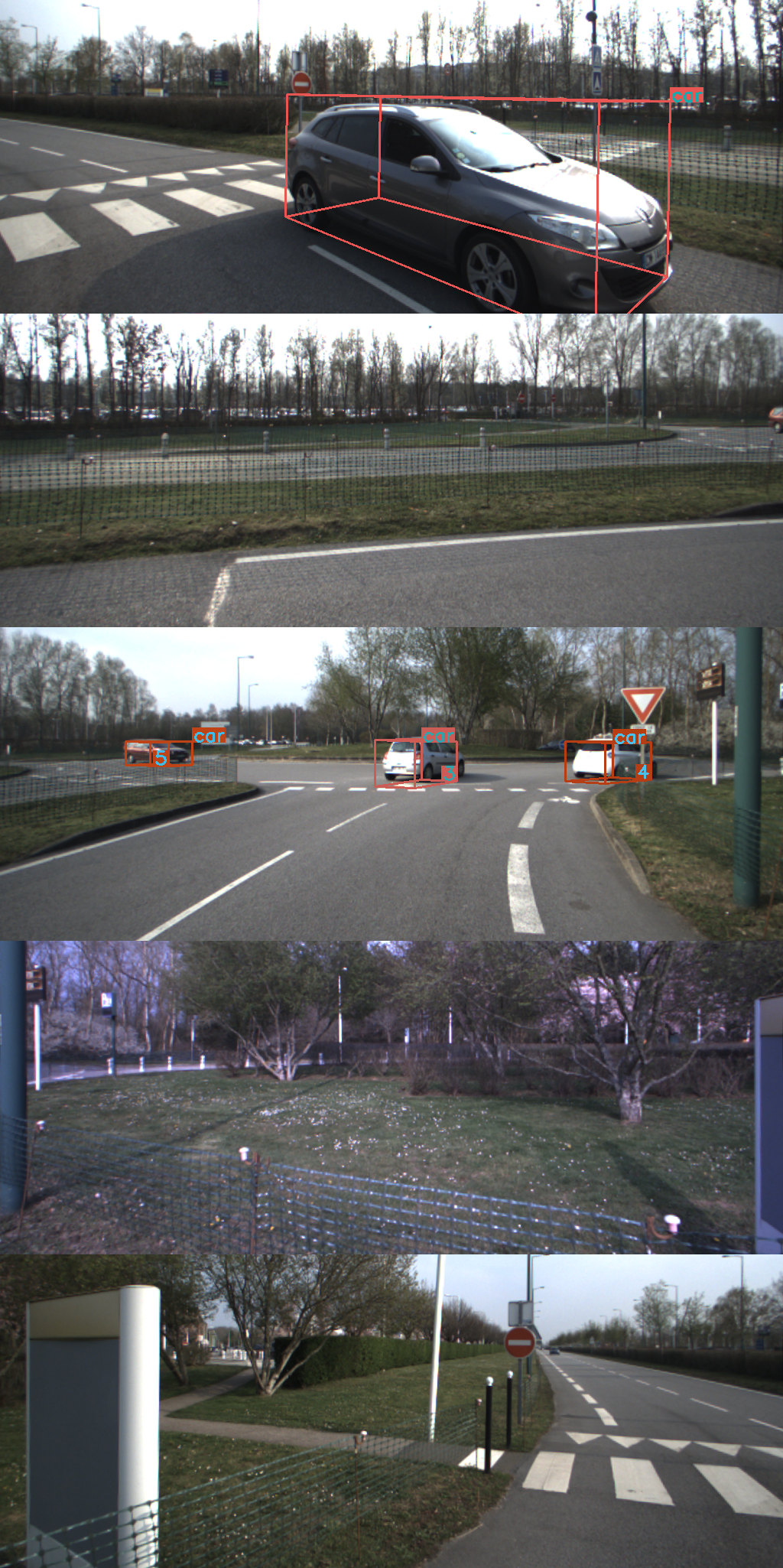}\hspace{0.7px}
    \includegraphics[width=0.16\linewidth]{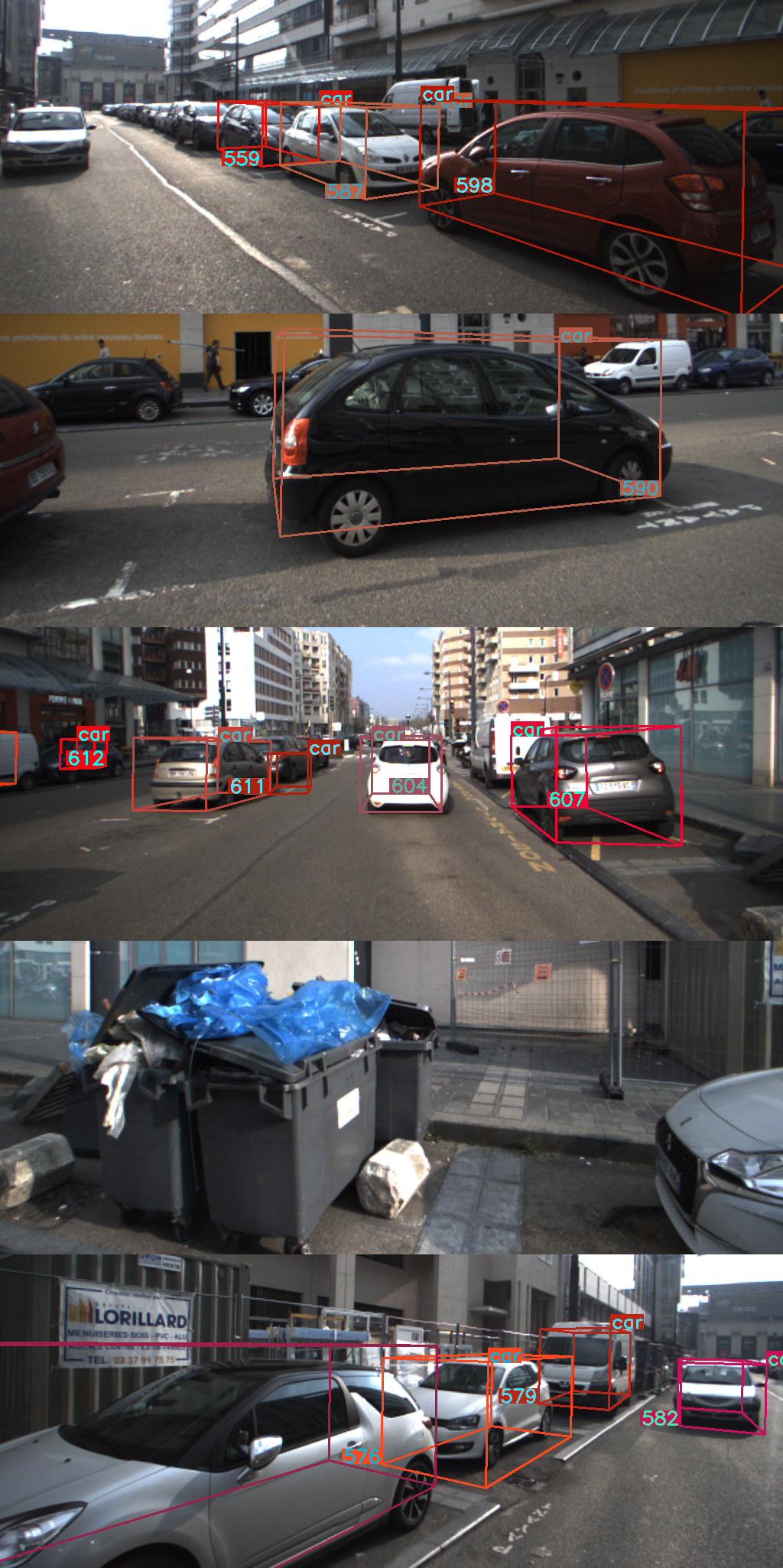}\hspace{0.7px}
    \includegraphics[width=0.16\linewidth]{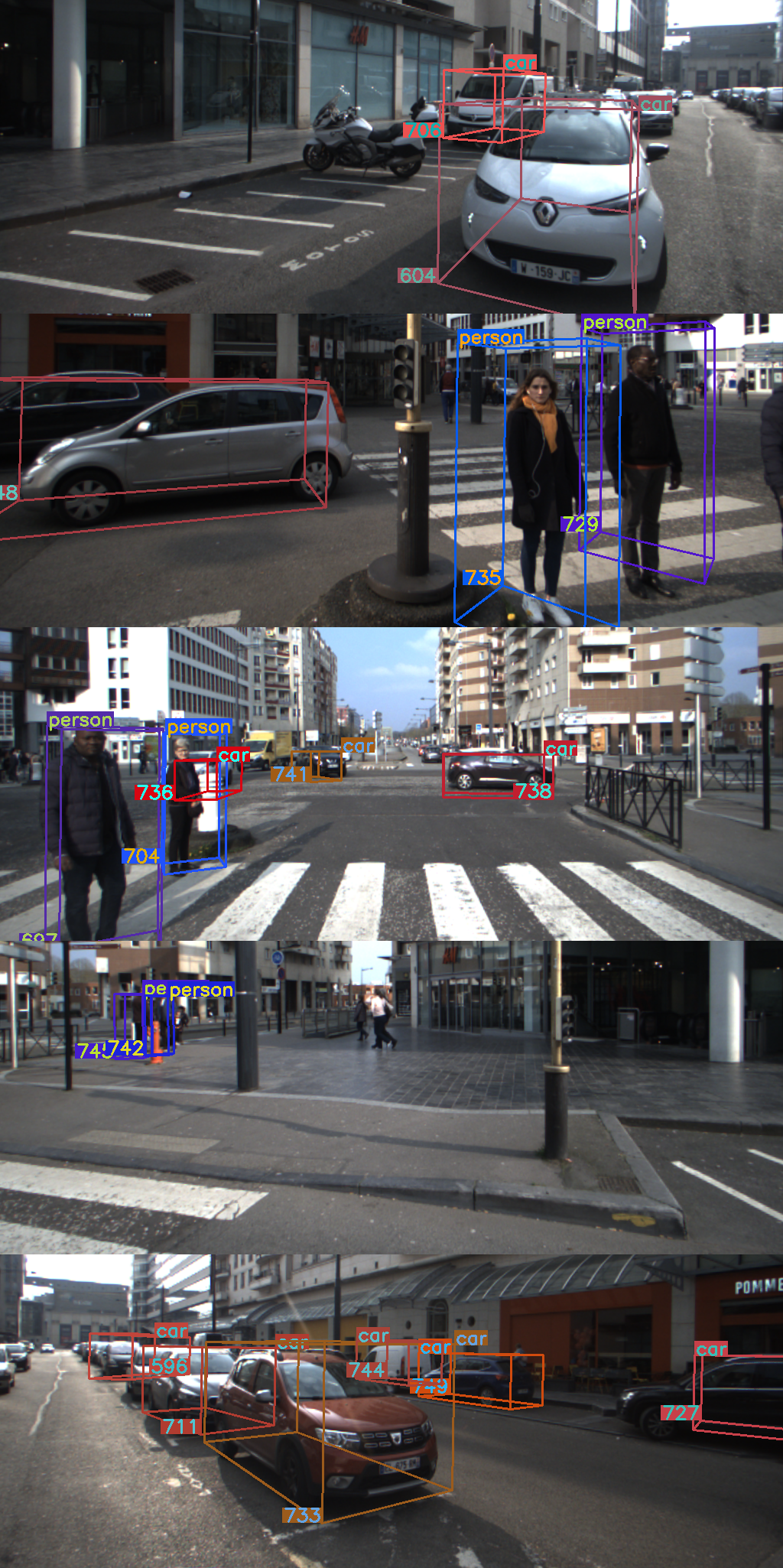}\hspace{0.7px}
    \includegraphics[width=0.16\linewidth]{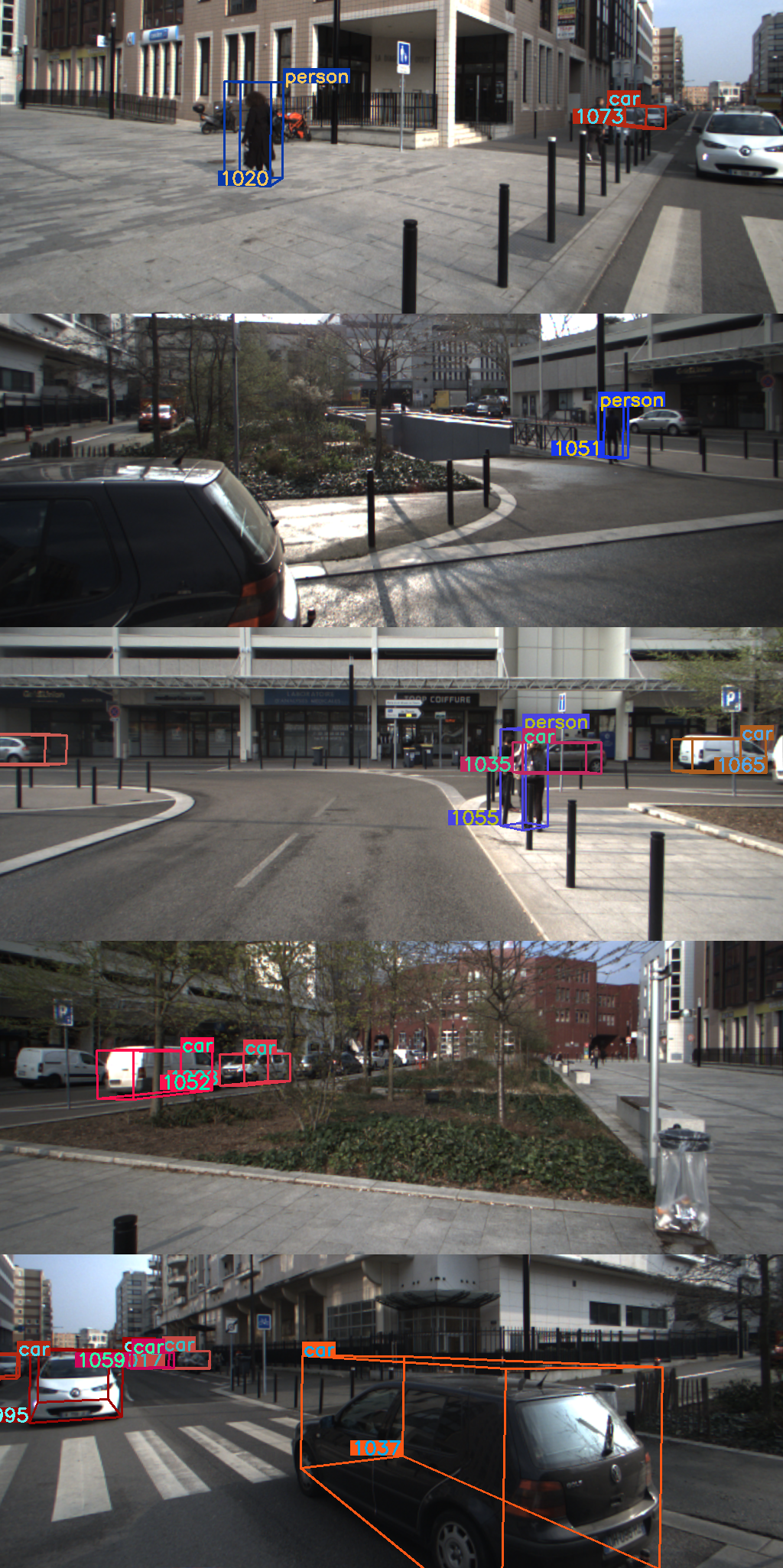}\hspace{0.7px}
    \includegraphics[width=0.16\linewidth]{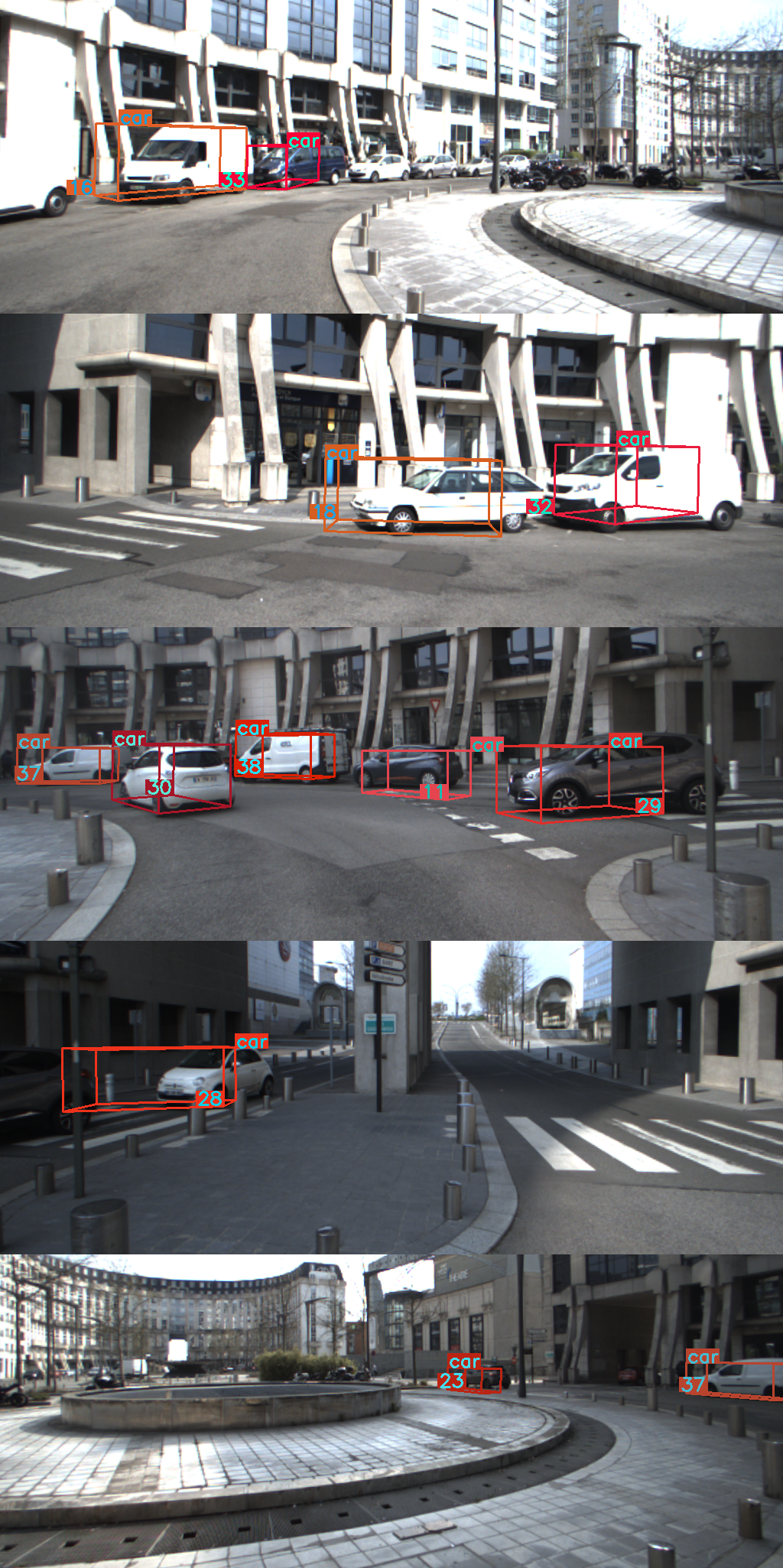}
} \\
\subfloat{
\\
    \includegraphics[width=0.16\linewidth]{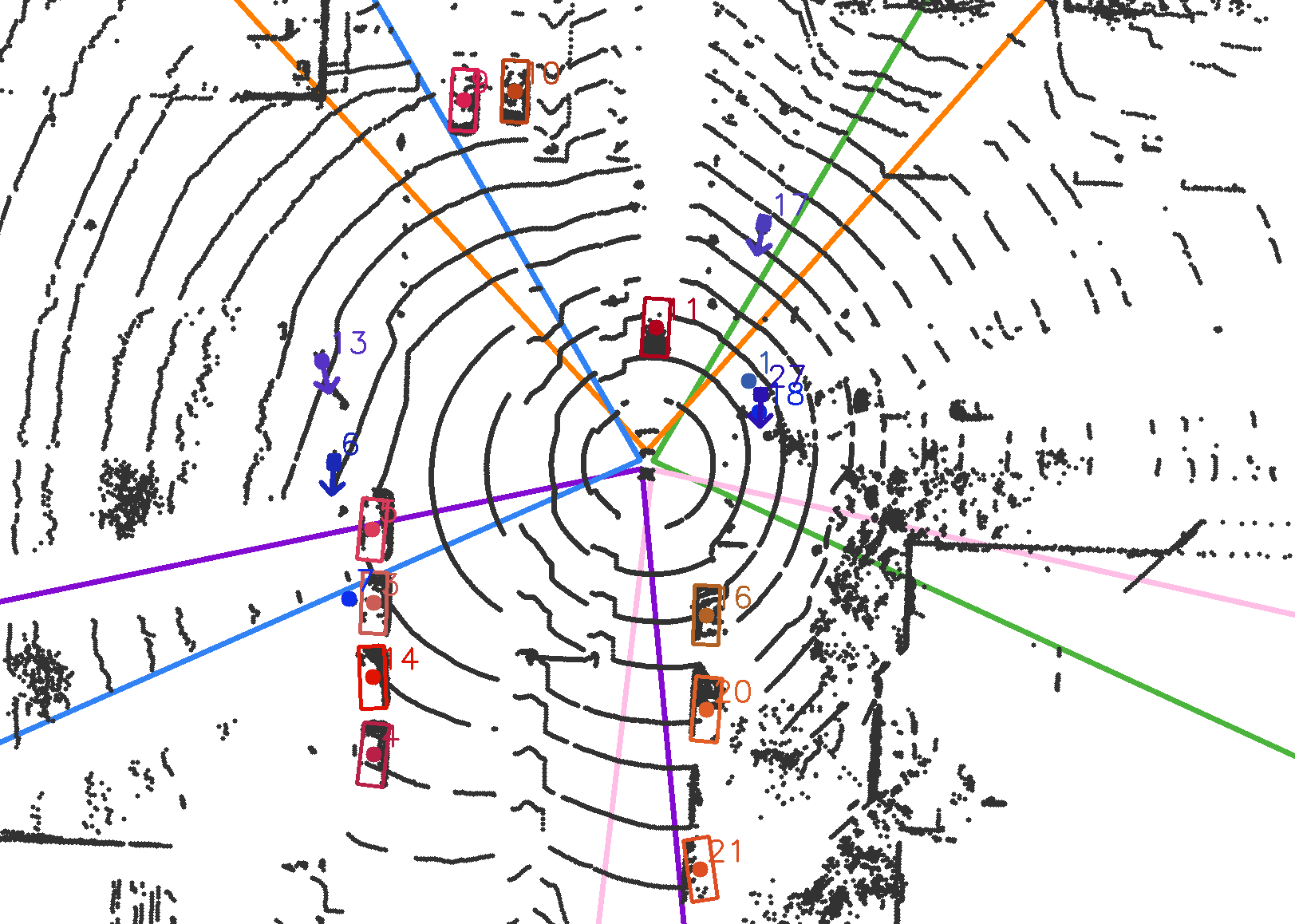}\hspace{0.7px}
    \includegraphics[width=0.16\linewidth]{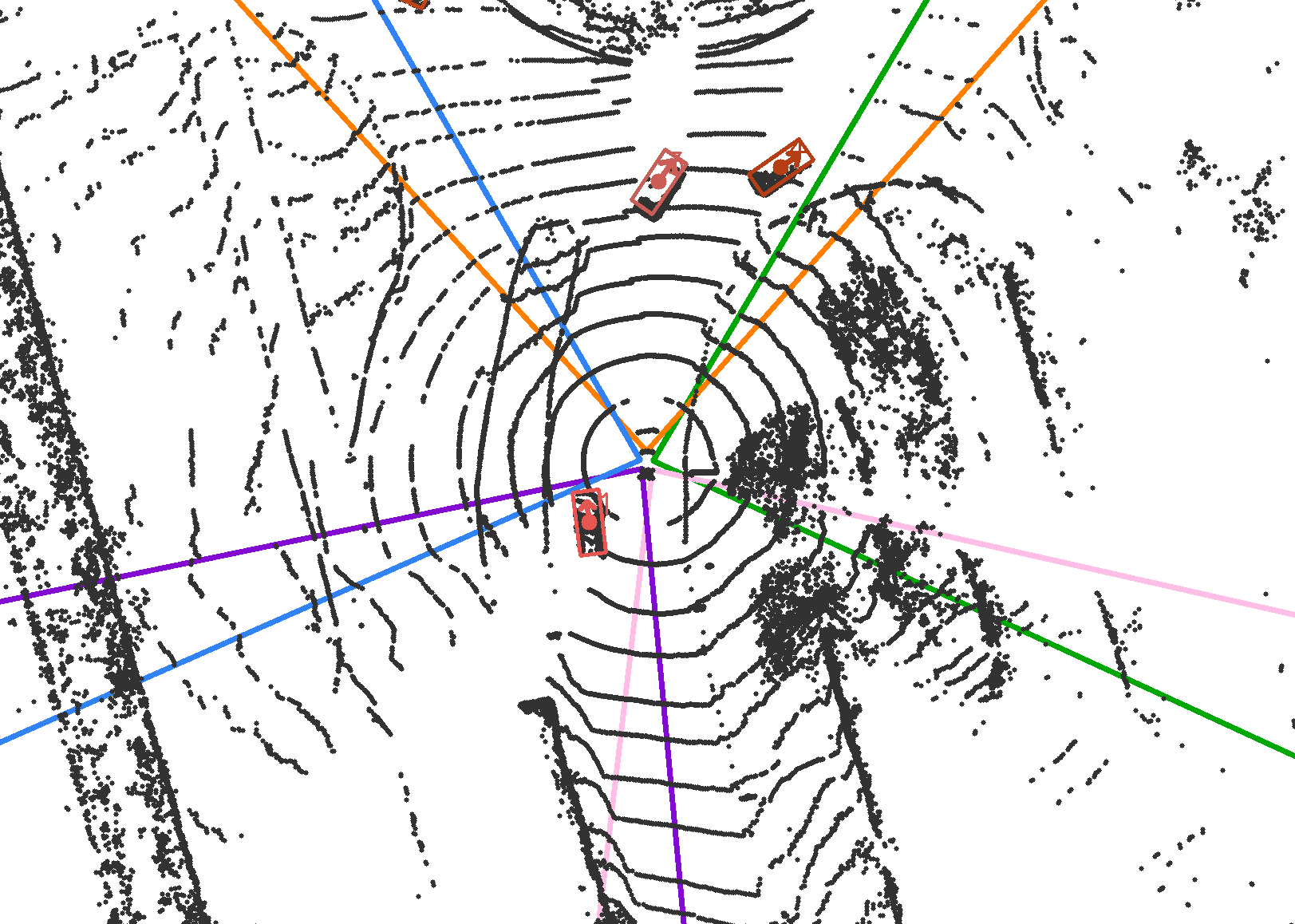}\hspace{0.7px}
    \includegraphics[width=0.16\linewidth]{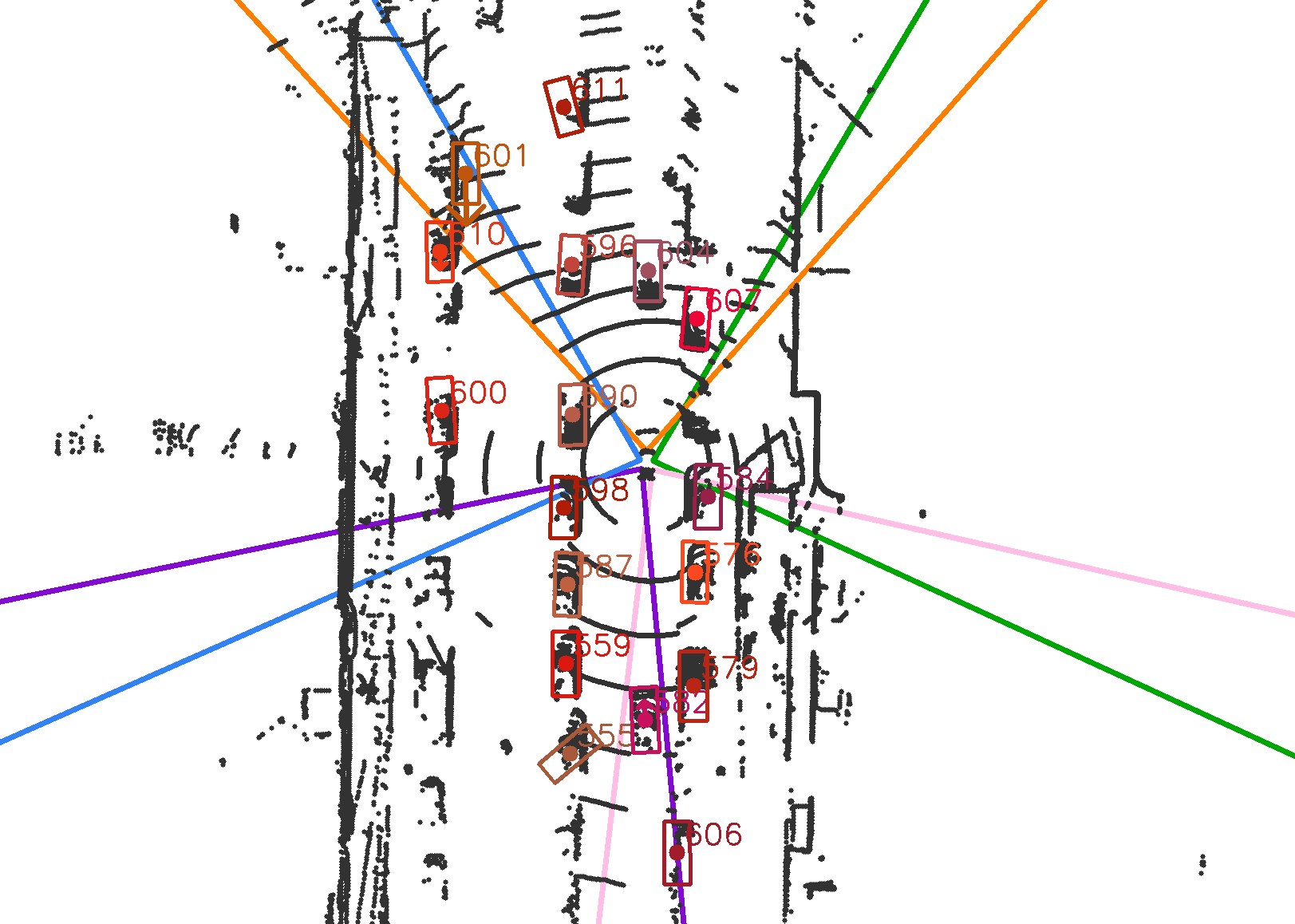}\hspace{0.7px}
    \includegraphics[width=0.16\linewidth]{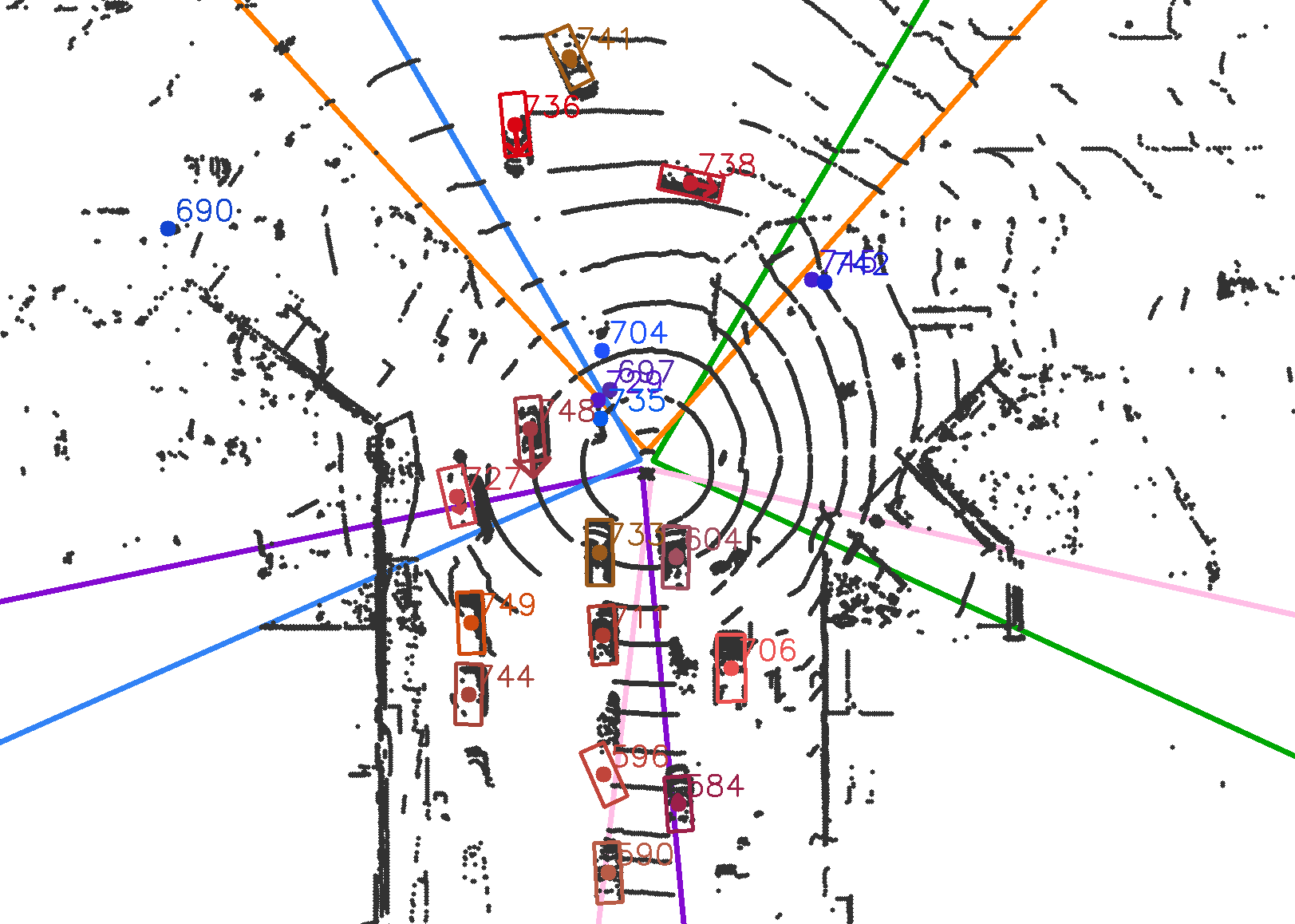}\hspace{0.7px}
    \includegraphics[width=0.16\linewidth]{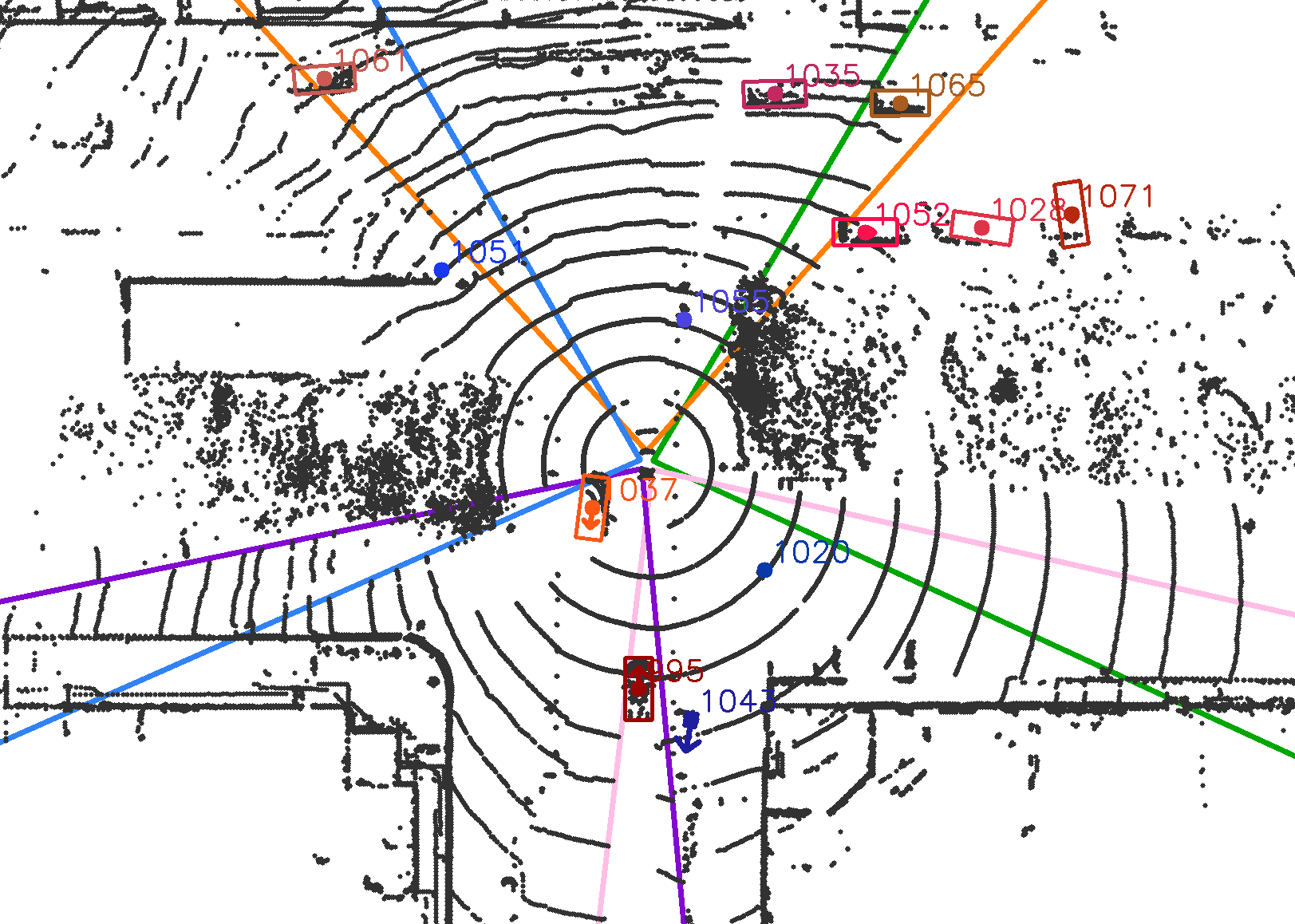}\hspace{0.7px}
    \includegraphics[width=0.16\linewidth]{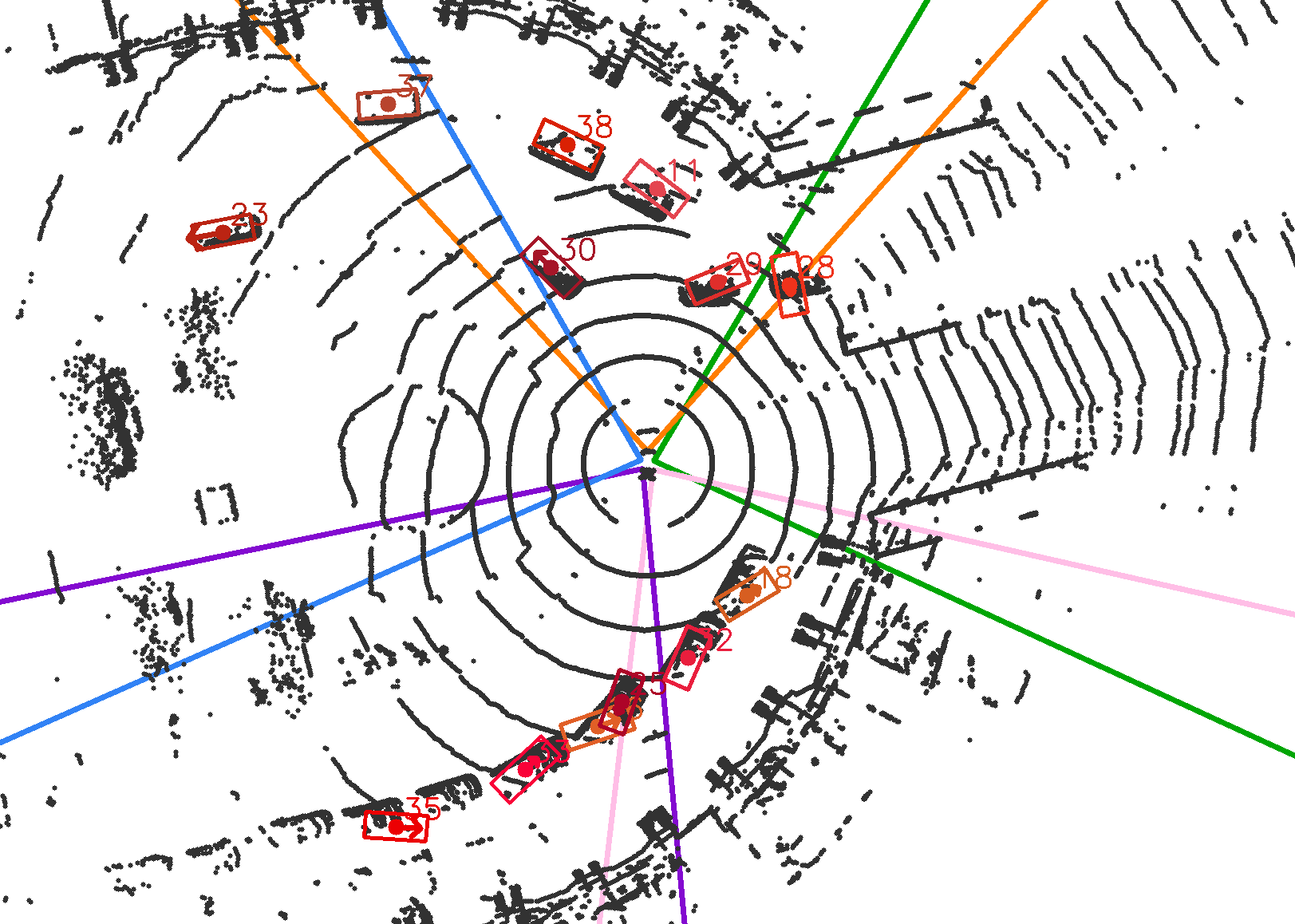}
}
\caption{Qualitative results of the proposed system on some typical traffic scenarios. From top to bottom: 3D detections in rear-left, front-left, front, front-right, and rear-right cameras, and Bird's Eye View representation.}
\label{fig:examples}
\end{figure*}

Fig.~\ref{fig:examples} depicts some examples of the performance of our detection solution in different traffic situations. As evidenced by these results, the 360\degree\ capabilities of the system make it suitable for the identification of the dynamic obstacles even in challenging situations such as intersections and roundabouts. Most roads users are correctly detected and provided with a representative 3D box. Some difficulties are found at the overlapping areas between consecutive cameras, where near objects appear truncated in both images leading to spurious 3D estimates. 

The system is able to output tracked agents at around \SI{10}{Hz}, which validates the viability of the approach for onboard perception. Further time optimization may be achieved by using dedicated hardware and software implementations.

\section{CONCLUSIONS}
The work presented in this paper provides a step forward in the development of a full 360\degree\ environment perception framework for autonomous driving applications. It takes advantage of the most advanced technologies of perception in a novel sensor fusion configuration that provides accurate and reliable detections.

The proposed solution has been validated in real traffic, with challenging scenarios where autonomous vehicles usually face difficulties, i.e., urban and peri-urban areas, and different visibility conditions. The tests have proved the suitability to handle complex situations, tracking many concurrent agents in real-time. 



Future works will aim to provide a solution to some of the limitations found in the current configuration. Hence, the addition of further classes would allow identifying a wider variety of agents involved in driving scenarios. Besides, issues related to incomplete LiDAR data may be solved by aggregating spatial information of objects detected in contiguous cameras as in \cite{Cortes2020}.


\section*{ACKNOWLEDGMENT}

Authors thank ANRT FUI Tornado project to partially fund this  work. They also want to express their gratitude to the Renault Research Department for its support in developing experimental  test. The work was also partially funded by the Spanish Government (TRA2016-78886-C3-1-R and RTI2018-096036-B-C21), the Universidad Carlos III of Madrid (PEAVAUTO-CM-UC3M) and the Comunidad de Madrid (SEGVAUTO-4.0-CM P2018/EMT-4362).

\bibliographystyle{IEEEtran}
\bibliography{good-library}

\end{document}